\documentclass{article}

\usepackage{iclr2027_conference,times}
\usepackage{amsmath,amssymb,mathtools}
\usepackage{amsthm}
\usepackage{booktabs}
\usepackage{longtable}
\usepackage{array}
\usepackage{graphicx}
\usepackage{xcolor}
\usepackage{tikz}
\usetikzlibrary{arrows.meta,backgrounds,calc,fit,positioning}
\usepackage{microtype}
\usepackage{enumitem}
\usepackage{float}
\usepackage{placeins}
\usepackage{wrapfig}
\makeatletter
\newcommand{\needlines}[1]{\par\penalty-100\begingroup\setlength{\dimen@}{#1}\dimen@ii\pagegoal\advance\dimen@ii-\pagetotal\ifdim\dimen@>\dimen@ii\ifdim\dimen@ii>\z@\vfil\fi\break\fi\endgroup}
\makeatother
\usepackage[T1]{fontenc}
\usepackage[utf8]{inputenc}
\usepackage{algorithm}
\usepackage{algpseudocode}
\usepackage{url}
\usepackage[colorlinks=true,allcolors=blue!45!black]{hyperref}
\hypersetup{pdfauthor={Xiaojing Du, Jiuyong Li, Lin Liu, Debo Cheng, Jixue Liu, Thuc Duy Le},pdftitle={Peer Effects in Signed Networks: Separating Influence Through Positive and Negative Ties}}
\ifdefined\pdfinfoomitdate \pdfinfoomitdate=1 \pdfsuppressptexinfo=-1 \pdftrailerid{} \fi
\algrenewcommand\algorithmicrequire{\textbf{Input:}}
\algrenewcommand\algorithmicensure{\textbf{Output:}}

\newcommand{\caseMainNPos}{6,375}

\newcommand{\caseMainNInt}{941}

\newcommand{\caseMainStudents}{11,117}

\newcommand{\timeFast}{11}

\newcommand{\timeSlow}{94}

\newcommand{\mvAblMeanBiasSide}{0.020}

\newcommand{\mvAblMeanSdSide}{0.084}
\newcommand{\mvAblMeanBiasOutcome}{0.054}

\newcommand{\mvAblMeanSdOutcome}{0.055}
\newcommand{\mvRepMain}{10}
\newcommand{\mvRepCoverage}{200}
\newcommand{\mvCovThreshold}{0.919}

\newcommand{\vtCaseBonfLoPos}{$-$0.3}
\newcommand{\vtCaseBonfHiPos}{6.9}

\newcommand{\vtCaseOutcomeMeanPct}{15.4}
\newcommand{\vtCaseUnsignedPP}{3.2}

\newcommand{\vcMinCov}{0.850}

\newcommand{\vbPosAlphaCov}{0.880}
\newcommand{\vbPosOtcCov}{0.840}

\newcommand{\vtCaseUnsignedSpendSharePct}{85}
\newcommand{\caseMainEstPosPP}{3.3}
\newcommand{\caseMainLoPosPP}{0.5}
\newcommand{\caseMainHiPosPP}{6.1}
\providecommand{\caseMainNPos}{6,375}
\newcommand{\caseMainEstNegPP}{-0.9}
\newcommand{\caseMainLoNegPP}{-7.2}
\newcommand{\caseMainHiNegPP}{5.5}

\providecommand{\caseMainNInt}{941}

\newcommand{\caseMainSchoolCount}{28}

\title{Peer Effects in Signed Networks:\\Separating Influence Through Positive and Negative Ties}

\author{\begin{minipage}{\dimexpr\textwidth-2\tabcolsep\relax}\centering\normalfont
\small\textbf{Xiaojing Du\textsuperscript{1}, Jiuyong Li\textsuperscript{1}, Lin Liu\textsuperscript{1}, Debo Cheng\textsuperscript{2}, Jixue Liu\textsuperscript{1}, Thuc Duy Le\textsuperscript{1}}\\[3pt]
\small\textsuperscript{1}Adelaide University, Adelaide, Australia\\
\small\textsuperscript{2}Hainan University, Haikou, China\\[3pt]
{\fontsize{7}{9}\selectfont\ttfamily xiaojing.du@adelaide.edu.au; jiuyong.li@adelaide.edu.au; lin.liu@adelaide.edu.au}\\
{\fontsize{7}{9}\selectfont\ttfamily jixue.liu@adelaide.edu.au; thuc.le@adelaide.edu.au; chengd@hainanu.edu.cn}
\end{minipage}}

\theoremstyle{plain}

\newtheorem{proposition}{Proposition}

\theoremstyle{definition}
\newtheorem{definition}{Definition}
\newcommand{\definitionrestatementname}{Definition}
\newtheorem*{definitionrestatement}{\definitionrestatementname}

\newtheorem{assumption}{Assumption}

\newcommand{\ind}{\mathrel{\perp\!\!\!\perp}}

\newcommand{\move}[2]{#1\!\to\!#2}

\iclrfinalcopy
\begin{document}
\raggedbottom
\setlength{\textfloatsep}{8pt plus 2pt minus 2pt}
\setlength{\floatsep}{8pt plus 2pt minus 2pt}
\maketitle
\lhead{Preprint}

\begin{abstract}
Evaluating network interventions requires understanding how treatment affects people through their social relationships. Counting treated neighbors without distinguishing supportive and antagonistic ties can conceal opposing influences. We define effects through positive and negative ties, their interaction, and a sign-composition effect of reallocating treatment between the two types at a fixed total, and give their identification formulas. Under sign-blind assignment, we show how ignoring signs mixes the effects of the two tie types. We propose SiDE (\textbf{Si}gned-exposure \textbf{D}oubly robust \textbf{E}stimator), which combines sign-specific outcome models with exposure probabilities induced by individual treatment assignment. We establish double robustness of its score and assess approximate intervals that account for overlapping neighborhoods. Semi-synthetic experiments on six real signed networks demonstrate accurate effect estimation and examine the limits of interval coverage. An exploratory reanalysis of published school-experiment data yields a positive estimate of the peer effect through spend-time ties on wristband wearing, but the intervals for all four effects include zero after adjustment for multiple comparisons. This framework can inform network intervention design by showing when influences through the two tie types reinforce or offset one another.
\end{abstract}

\section{Introduction}
\label{sec:intro}

Social networks record trust and distrust on online platforms \citep{guha2004,leskovec2010} and friendship and antagonism in villages \citep{ghasemian2024}. Signed networks represent these relationships with positive ($+$) and negative ($-$) ties \citep{cartwright1956,tang2016}. A \emph{peer effect} is a change in a person's outcome caused by changing peers' treatments while holding that person's own treatment fixed \citep{hudgens2008}. This dependence on others' treatments is network interference. Peer effects arise in contexts such as village health programs \citep{kim2015}, online experiments \citep{eckles2017}, and school interventions \citep{paluck2016}.

\begin{figure}[t]
\centering
\begin{tikzpicture}[
    scale=0.68, every node/.style={font=\small},
    ego/.style={circle,draw=blue!55!black,fill=blue!8,very thick,minimum size=6.4mm,inner sep=0pt},
    treated/.style={circle,draw=black,fill=black!75,minimum size=4.8mm,inner sep=0pt},
    untreated/.style={circle,draw=black,fill=white,minimum size=4.8mm,inner sep=0pt},
    posline/.style={draw=teal!70!black,very thick},
    negline/.style={draw=red!65!black,very thick,dashed}
]
\begin{scope}[shift={(0,0)}]
  \node[ego] (a) at (0,0) {A};
  \node[treated]   (a1) at (-1.75,1.05) {};
  \node[treated]   (a2) at (-0.60,1.75) {};
  \node[treated]   (a3) at ( 0.60,1.75) {};
  \node[untreated] (a4) at ( 1.75,1.05) {};
  \node[treated]   (a5) at (-1.35,-1.30) {};
  \node[untreated] (a6) at ( 1.35,-1.30) {};
  \draw[posline] (a)--(a1); \draw[posline] (a)--(a2); \draw[posline] (a)--(a3); \draw[posline] (a)--(a4);
  \draw[negline] (a)--(a5); \draw[negline] (a)--(a6);
  \node[align=center,font=\scriptsize] at (0,-2.15) {Treated neighbors:\\3 trusted and 1 avoided};
\end{scope}
\begin{scope}[shift={(4.7,0)}]
  \node[ego] (b) at (0,0) {B};
  \node[treated]   (b1) at (-1.75,1.05) {};
  \node[treated]   (b2) at (-0.60,1.75) {};
  \node[treated]   (b3) at ( 0.60,1.75) {};
  \node[treated]   (b4) at ( 1.75,1.05) {};
  \node[untreated] (b5) at (-1.35,-1.30) {};
  \node[untreated] (b6) at ( 1.35,-1.30) {};
  \draw[posline] (b)--(b1); \draw[posline] (b)--(b2); \draw[posline] (b)--(b3); \draw[posline] (b)--(b4);
  \draw[negline] (b)--(b5); \draw[negline] (b)--(b6);
  \node[align=center,font=\scriptsize] at (0,-2.15) {Treated neighbors:\\4 trusted and 0 avoided};
\end{scope}
\node[font=\footnotesize] at (2.35,2.55) {(a) Signed neighborhoods};
\begin{scope}[shift={(-1.55,-3.05)}]
  \draw[posline] (0,0)--(0.75,0); \node[font=\scriptsize,anchor=west] at (0.80,0) {trusted};
  \draw[negline] (2.55,0)--(3.30,0); \node[font=\scriptsize,anchor=west] at (3.35,0) {avoided};
  \node[treated,minimum size=2.6mm] at (5.35,0) {}; \node[font=\scriptsize,anchor=west] at (5.55,0) {treated};
  \node[untreated,minimum size=2.6mm] at (7.30,0) {}; \node[font=\scriptsize,anchor=west] at (7.50,0) {untreated};
\end{scope}
\begin{scope}[shift={(9.8,-1.55)}, x=1.1cm, y=1.1cm]
  \fill[orange!30] (4,0) rectangle (5,1);
  \fill[orange!30] (3,1) rectangle (4,2);
  \draw[step=1,black!40] (0,0) grid (5,3);
  \draw[black] (0,0) rectangle (5,3);
  \draw[-{Latex[length=1.4mm]},black!70] (0,-0.62) -- (5.35,-0.62);
  \draw[-{Latex[length=1.4mm]},black!70] (-0.62,0) -- (-0.62,3.15);
  \foreach \x in {0,...,4} { \node[font=\scriptsize] at (\x+0.5,-0.32) {\x}; }
  \foreach \y in {0,...,2} { \node[font=\scriptsize] at (-0.32,\y+0.5) {\y}; }
  \node[font=\scriptsize] at (2.5,-0.85) {treated trusted (of 4)};
  \node[font=\scriptsize,rotate=90] at (-0.9,1.5) {treated avoided (of 2)};
  \node[circle,draw=blue!55!black,fill=blue!8,thick,minimum size=4.2mm,inner sep=0pt,font=\scriptsize] at (3.5,1.5) {A};
  \node[circle,draw=blue!55!black,fill=blue!8,thick,minimum size=4.2mm,inner sep=0pt,font=\scriptsize] at (4.5,0.5) {B};
  \draw[-{Latex[length=1.6mm]},teal!70!black,very thick] (0.6,0.5) -- (1.45,0.5);
  \draw[-{Latex[length=1.6mm]},red!65!black,very thick] (0.5,0.6) -- (0.5,1.45);
  \draw[-{Latex[length=1.6mm]},teal!70!black,very thick,densely dotted] (0.6,1.5) -- (1.45,1.5);
  \draw[-{Latex[length=1.6mm]},orange!80!black,very thick] (3.76,1.24) -- (4.24,0.76);
  \node[font=\footnotesize] at (2.5,3.73) {(b) Signed exposure comparisons};
  \node[align=left,font=\scriptsize,anchor=north west] at (-1.15,-1.15) {%
    $\tau_{+}$: \textcolor{teal!70!black}{$\boldsymbol\rightarrow$} positive-tie increment\\
    $\tau_{-}$: \textcolor{red!65!black}{$\boldsymbol\uparrow$} negative-tie increment};
  \node[align=left,font=\scriptsize,anchor=north west] at (3.20,-1.15) {%
    $\tau_{\mathrm{int}}$: \tikz[baseline=-0.55ex]\draw[-{Latex[length=1.2mm]},teal!70!black,very thick,densely dotted] (0,0)--(0.38,0); $-$ \tikz[baseline=-0.55ex]\draw[-{Latex[length=1.2mm]},teal!70!black,very thick] (0,0)--(0.38,0);\\
    $\kappa$: \textcolor{orange!80!black}{$\mathrm{A}\!\rightarrow\!\mathrm{B}$}, same total};
\end{scope}
\end{tikzpicture}
\caption{Exposure patterns with four treated neighbors. The grid counts treated trusted neighbors horizontally and treated avoided neighbors vertically. Mean outcome changes along horizontal and vertical arrows define the positive- and negative-tie effects. Their interaction subtracts the solid horizontal increment from the dotted one. The sign-composition effect compares outcomes under exposures A=$(3,1)$ and B=$(4,0)$.}
\label{fig:motivation}
\end{figure}
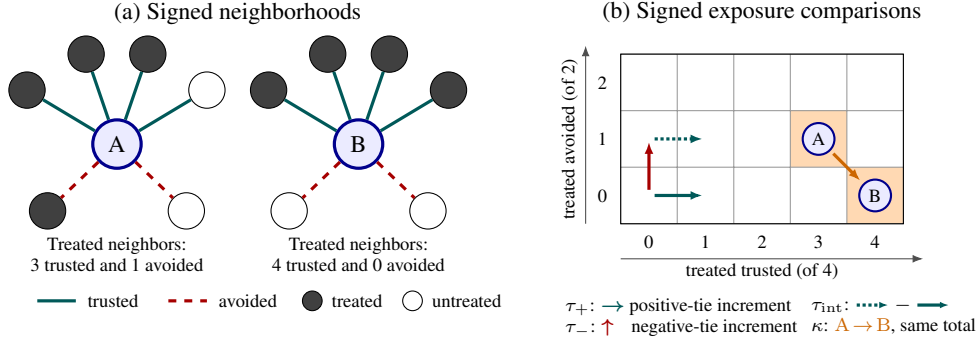

Consider a village health program in which treatment is receipt of a preventive product through the program and the outcome is subsequent use. In Figure~\ref{fig:motivation}(a), households A and B each have four trusted and two avoided neighbors. A's treated neighbors are three trusted households and one avoided household, and B's are four trusted households. Figure~\ref{fig:motivation}(b) plots these signed exposures as $(3,1)$ and $(4,0)$. With a household's own treatment fixed, the \emph{positive-tie effect} and \emph{negative-tie effect} measure the response to treating an additional trusted or avoided neighbor, respectively, keeping the other treated-neighbor count fixed. Both effects can be positive or negative. Their \emph{interaction} asks whether treating avoided neighbors changes the effect of treating trusted neighbors. The \emph{sign-composition effect} compares a household's responses to A's and B's exposure patterns: treatment shifts from an avoided to a trusted neighbor while four neighbors remain treated.

An analysis that counts treated neighbors without distinguishing tie signs assigns A and B the same exposure. An estimated effect near zero can then reflect weak peer influence or opposing influences that cancel out. Existing frameworks for causal inference under interference accommodate exposure mappings and contrasts between exposure states \citep{aronow2017,qu2024}. Regression and graph neural estimators adjust for individual and neighborhood covariates \citep{forastiere2021,ma2021,jiang2022,zhao2024,leung2025}. Building on these frameworks, we study effects through positive and negative ties, including the reallocation of treatment between peer types at a fixed total, and characterize the information lost when tie signs are discarded.

To construct a doubly robust score for these contrasts, we model both conditional outcomes and the joint probabilities of own treatment and the two signed exposure counts. Overlapping neighborhoods can also induce dependence between units, which must be considered when assessing uncertainty. We propose SiDE (\textbf{Si}gned-exposure \textbf{D}oubly robust \textbf{E}stimator), which combines separate encoders for positive and negative neighborhoods with exposure probabilities induced by individual treatment assignment. We also evaluate approximate intervals that account for dependence between units with overlapping neighborhoods.

\textbf{Contributions.} (1) We define and identify four signed peer effects: the positive-tie effect, the negative-tie effect, their interaction, and the sign-composition effect. Under sign-blind assignment, we derive a closed-form decomposition of the unsigned increment as a weighted average of positive- and negative-tie increments, with weights equal to the proportions of the two tie types. This decomposition shows when opposing increments cancel out. (2) We propose SiDE, establish the double robustness of its score, and construct approximate intervals that account for overlapping neighborhoods. (3) We evaluate SiDE against eleven baselines on semi-synthetic data built on six real signed networks and reanalyze data from an existing school field experiment to separate peer effects through students' spend-time and conflict ties.

\section{Related Work}
\label{sec:related}

\textbf{Causal inference under network interference.} Potential outcomes describe an ego's outcome under specified treatments of the ego and its peers \citep{hudgens2008}. An exposure mapping summarizes peers' assignments, for example by the number of treated neighbors \citep{aronow2017}. Observational analyses adjust for ego and neighbor covariates \citep{forastiere2021}. Targeted learning updates outcome predictions using assignment probabilities \citep{ogburn2024}, while graph neural models learn adjustments from neighborhood information \citep{ma2021,jiang2022,zhao2024,adhikari2025,caljon2026}. Doubly robust and AIPW estimators combine outcome and assignment models, with network-dependent variance estimates \citep{leung2025} or type-specific treated-neighbor counts \citep{qu2024}. Studies also distinguish tie strength \citep{bond2012,rajkumar2022}, edge type \citep{lin2023}, and multiple networks \citep{egami2021}. Our work builds on these foundations to estimate signed peer effects and to characterize how ignoring tie signs combines positive- and negative-tie increments under sign-blind assignment.

\textbf{Signed networks.} Positive and negative ties enter models of structural balance and social status \citep{cartwright1956,leskovec2010b}, trust propagation \citep{guha2004}, and network representation learning \citep{derr2018,tang2016}. Models of conflict and diffusion examine how allies and enemies affect behavior \citep{konig2017,li2013}. Empirical studies examine social influence in signed networks. \citet{ghasemian2024} study information diffusion through antagonistic ties, while \citet{bellotti2026} use cross-sectional data and a calibrated behavioral model to examine how negative ties may weaken reinforcement from positive contacts. Disliking a source need not reverse its influence \citep{takacs2016}. These findings lead us to define effects through each tie type, the interaction between the two types, and the reallocation of treatment between them at a fixed total.

\section{Problem Setting}

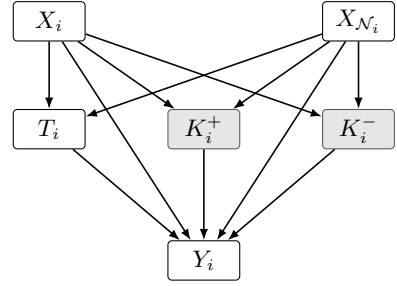
\begin{wrapfigure}{r}{0.47\textwidth}
\vspace{-0.9em}
\centering
\begin{tikzpicture}[
    scale=0.62,
    every node/.style={font=\footnotesize},
    obs/.style={draw=black,rounded corners=1.5pt,minimum width=9.5mm,minimum height=5.2mm,inner sep=1.5pt},
    det/.style={draw=black!65,fill=black!10,rounded corners=1.5pt,minimum width=9.5mm,minimum height=5.2mm,inner sep=1.5pt},
    arr/.style={-{Latex[length=1.5mm]},semithick}
]
    \node[obs] (xi) at (-3.3, 2.50) {$X_i$};
    \node[obs] (xn) at ( 3.3, 2.50) {$X_{\mathcal N_i}$};
    \node[obs] (ti) at (-3.3, 0.20) {$T_i$};
    \node[det] (kp) at ( 0.0, 0.20) {$K_i^{+}$};
    \node[det] (km) at ( 3.3, 0.20) {$K_i^{-}$};
    \node[obs] (yi) at ( 0.0,-2.60) {$Y_i$};
    \draw[arr] (xi) -- (ti);
    \draw[arr] (xn) -- (ti);
    \draw[arr] (xi) -- (kp);
    \draw[arr] (xi) -- (km);
    \draw[arr] (xn) -- (kp);
    \draw[arr] (xn) -- (km);
    \draw[arr] (ti) -- (yi);
    \draw[arr] (kp) -- (yi);
    \draw[arr] (km) -- (yi);
    \draw[arr] (xi) -- (yi);
    \draw[arr] (xn) -- (yi);
\end{tikzpicture}
\caption{Causal graph for one ego. Shaded nodes are the numbers of treated neighbors within $\mathcal N_i^{+}$ and within $\mathcal N_i^{-}$.}
\label{fig:dag}
\vspace{-0.6em}
\end{wrapfigure}

Let $V=\{1,\ldots,n\}$ index the units in a directed network observed before treatment. The signed adjacency matrix $S=(S_{ij})_{i,j\in V}$ has entries $S_{ij}\in\{-1,0,+1\}$ recording a negative tie, no tie, or a positive tie from unit $i$ to unit $j$, with $S_{ii}=0$. The two neighborhoods are $\mathcal N_i^{+}=\{j:S_{ij}=+1\}$ and $\mathcal N_i^{-}=\{j:S_{ij}=-1\}$, with degrees $d_i^{+}=|\mathcal N_i^{+}|$ and $d_i^{-}=|\mathcal N_i^{-}|$ and union $\mathcal N_i$. Each unit is an ego when its outcome is studied and a peer when it belongs to another unit's neighborhood. A peer effect runs from $j$'s treatment to $i$'s outcome.

Let $T_i\in\{0,1\}$ be unit $i$'s own treatment, $Y_i$ the observed outcome, and $X_i$ a vector of pre-treatment characteristics and network features. Write $\mathbf T=(T_1,\ldots,T_n)$, $\mathbf X=(X_1,\ldots,X_n)$, and $X_{\mathcal N_i}=(\{X_j\}_{j\in\mathcal N_i^{+}},\{X_j\}_{j\in\mathcal N_i^{-}})$, keeping neighbor covariates separate by sign and invariant to their order. The signed exposures are $K_i^{+}(\mathbf T)=\sum_{j\in\mathcal N_i^{+}}T_j$ and $K_i^{-}(\mathbf T)=\sum_{j\in\mathcal N_i^{-}}T_j$. Together with $T_i$ they form the joint state $Z_i=(T_i,K_i^{+},K_i^{-})$. Lowercase $k^{+}$ and $k^{-}$ denote specified nonnegative integer values of these exposure counts. An unsigned analysis retains only $K_i^{u}=K_i^{+}+K_i^{-}$ and cannot distinguish A's exposure $(3,1)$ from B's $(4,0)$ in Figure~\ref{fig:motivation}. Figure~\ref{fig:dag} shows the causal graph.

\subsection{Assumptions}
\label{sec:assumptions}

Let $Y_i(\mathbf t)$ denote the potential outcome under assignment $\mathbf t$, and $Y_i$ the outcome observed under $\mathbf T$. Expectations average over a uniformly drawn ego and over covariates, assignment, and potential outcomes given the fixed network. A nonempty set $\mathcal U\subseteq V$, defined by signed degrees, specifies the target population, and conditioning on $i\in\mathcal U$ restricts the average to it. We use four identification conditions \citep{hudgens2008,aronow2017,forastiere2021}.

\begin{assumption}[Signed exposure mapping]
\label{as:exposure}
A neighbor's treatment affects ego $i$ only through $K_i^{+}(\mathbf t)$ and $K_i^{-}(\mathbf t)$, the numbers of treated positive- and negative-tie neighbors, so that $Y_i(\mathbf t)=Y_i\!\left(t_i,K_i^{+}(\mathbf t),K_i^{-}(\mathbf t)\right)$ for every assignment $\mathbf t$.
\end{assumption}

Under Assumption~\ref{as:exposure} we write $Y_i(t,k^{+},k^{-})$ for the potential outcome under own treatment $t$, $k^{+}$ treated positive-tie peers, and $k^{-}$ treated negative-tie peers.

\begin{assumption}[Consistency]
\label{as:consistency}
The observed outcome is the potential outcome at the realized exposure, $Y_i=Y_i\!\left(T_i,K_i^{+},K_i^{-}\right)$.
\end{assumption}

\begin{assumption}[Conditional exchangeability]
\label{as:exchange}
For every $t\in\{0,1\}$, $k^{+}\le d_i^{+}$, and $k^{-}\le d_i^{-}$, $Y_i(t,k^{+},k^{-})\ind (T_i,K_i^{+},K_i^{-})\mid X_i,X_{\mathcal N_i}$. Conditional on ego and neighbor covariates, the joint state carries no further information about potential outcomes.

\end{assumption}

\begin{assumption}[Positivity]
\label{as:positivity}
Every state used in a reported contrast has positive conditional probability, $\Pr(T_i=t,K_i^{+}=k^{+},K_i^{-}=k^{-}\mid X_i,X_{\mathcal N_i})>0$, at almost every covariate value in its target population.

\end{assumption}

\subsection{The four effects}
\label{sec:estimands}

We now formalize the four effects introduced in Section~\ref{sec:intro}. Treated-neighbor counts cannot exceed the corresponding degrees, giving the reachable set $\mathcal V(k^{+},k^{-})=\{i:d_i^{+}\ge k^{+},\ d_i^{-}\ge k^{-}\}$. Each effect compares potential outcomes over a common nonempty population that can reach every compared state.

\begin{definition}[Positive-tie peer effect]
\label{def:taupos}
For own treatment $t$, negative-tie exposure $k^{-}$, and positive-tie levels $k_0^{+}<k_1^{+}$,
\begin{equation}
\label{eq:taupos}
\tau_{+}\!\left(\move{k_0^{+}}{k_1^{+}};\,t,k^{-}\right)
=
\mathbb E\!\left[\,Y_i(t,k_1^{+},k^{-})-Y_i(t,k_0^{+},k^{-})
\;\middle|\; i\in\mathcal V(k_1^{+},k^{-})\right].
\end{equation}
\end{definition}

This measures the effect of increasing positive-tie exposure, holding own treatment and negative-tie exposure fixed. The negative-tie effect reverses the roles of the two exposures.

\begin{definition}[Negative-tie peer effect]
\label{def:tauneg}
For own treatment $t$, positive-tie exposure $k^{+}$, and negative-tie levels $k_0^{-}<k_1^{-}$,
\begin{equation}
\label{eq:tauneg}
\tau_{-}\!\left(\move{k_0^{-}}{k_1^{-}};\,t,k^{+}\right)
=
\mathbb E\!\left[\,Y_i(t,k^{+},k_1^{-})-Y_i(t,k^{+},k_0^{-})
\;\middle|\; i\in\mathcal V(k^{+},k_1^{-})\right].
\end{equation}
\end{definition}

The subscripts $+$ and $-$ in $\tau_{+}$ and $\tau_{-}$ name the two tie types. Either effect can be positive, zero, or negative. Their interaction measures whether one exposure changes the effect of the other.

\begin{definition}[Interaction between tie types]
\label{def:tauint}
For own treatment $t$, positive-tie levels $k_0^{+}<k_1^{+}$, and negative-tie levels $k_0^{-}<k_1^{-}$,
\begin{equation}
\label{eq:tauint}
\begin{split}
\tau_{\mathrm{int}}\!\left(\move{k_0^{+}}{k_1^{+}},\,\move{k_0^{-}}{k_1^{-}};\,t\right)
=\mathbb E\Big[\,
&\big\{Y_i(t,k_1^{+},k_1^{-})-Y_i(t,k_0^{+},k_1^{-})\big\}\\
-\,&\big\{Y_i(t,k_1^{+},k_0^{-})-Y_i(t,k_0^{+},k_0^{-})\big\}
\;\Big|\; i\in\mathcal V(k_1^{+},k_1^{-})\Big].
\end{split}
\end{equation}
\end{definition}

A negative value means the positive-tie increment is smaller when negative-tie exposure is higher.

\begin{definition}[Sign-composition effect]
\label{def:kappa}
For own treatment $t$ and levels $k^{+}\ge0$ and $k^{-}\ge1$,
\begin{equation}
\label{eq:kappa}
\kappa\!\left(k^{+},k^{-};\,t\right)
=
\mathbb E\!\left[\,Y_i(t,k^{+}+1,k^{-}-1)-Y_i(t,k^{+},k^{-})
\;\middle|\; i\in\mathcal V(k^{+}+1,k^{-})\right].
\end{equation}
\end{definition}

This contrast measures the effect of moving treatment from a negative-tie to a positive-tie neighbor at a fixed total, as from A to B in Figure~\ref{fig:motivation}. Both endpoints are reachable on $\mathcal V(k^{+}+1,k^{-})$. There it equals the positive-tie increment minus the negative-tie increment from $(t,k^{+},k^{-}-1)$. The treated-neighbor total alone cannot express this comparison.

\subsection{Identification}
\label{sec:identification}

Let $Q_i=(X_i,X_{\mathcal N_i})$ collect ego $i$'s adjustment covariates, and define the observed outcome regression by
\begin{equation}
\label{eq:m}
m(t,k^{+},k^{-},q)=\mathbb E[Y_i\mid T_i=t,K_i^{+}=k^{+},K_i^{-}=k^{-},Q_i=q].
\end{equation}
\begin{proposition}[Identification]
\label{prop:id}
Under Assumptions~\ref{as:exposure}--\ref{as:positivity}, for any state $(t,k^{+},k^{-})$ and any nonempty degree-defined set $\mathcal U\subseteq\mathcal V(k^{+},k^{-})$ on which this state satisfies Assumption~\ref{as:positivity},
\begin{equation}
\label{eq:id}
\mathbb E[Y_i(t,k^{+},k^{-})\mid i\in\mathcal U]
=\mathbb E[m(t,k^{+},k^{-},Q_i)\mid i\in\mathcal U].
\end{equation}
\end{proposition}
Conditional exchangeability and consistency equate the conditional means of potential and observed outcomes, and averaging over $Q_i$ gives Eq.~\eqref{eq:id}. The four identified effects follow by replacing each potential outcome in Definitions~\ref{def:taupos}--\ref{def:kappa} by $m$ over the stated common population.

\subsection{What an unsigned analysis estimates}
\label{sec:aggregate}

Fix a population with signed degrees $(a,b)$, where $a,b\ge1$ and $N=a+b$, and write $\mu(t,k^{+},k^{-})=\mathbb E[Y_i(t,k^{+},k^{-})\mid d_i^{+}=a,d_i^{-}=b]$. An unsigned analysis adjusts for the same covariates but retains only the total $K_i^u$, targeting
\begin{equation}
\label{eq:muu}
\mu^{u}(t,k)=\mathbb E\big[\mathbb E[Y_i\mid T_i=t,K_i^{u}=k,Q_i]\mid d_i^{+}=a,d_i^{-}=b\big].
\end{equation}
\begin{proposition}[The unsigned peer effect combines the two tie types]
\label{prop:mixture}
Under Assumptions~\ref{as:exposure} to \ref{as:positivity} and sign-blind assignment (uniform neighbor subsets conditional on own treatment, total exposure, and covariates), for $0\le k<N$ such that $(T_i,K_i^{u})=(t,k)$ and $(t,k+1)$ both have positive probability given $Q_i$,
\begin{equation}
\label{eq:mixture}
\mu^{u}(t,k+1)-\mu^{u}(t,k)
=
\frac{a}{N}\,\delta^{+}(t,k)+\frac{b}{N}\,\delta^{-}(t,k),
\end{equation}
where $\delta^{+}(t,k)=\mathbb E[\mu(t,R_k^{+}+1,k-R_k^{+})-\mu(t,R_k^{+},k-R_k^{+})]$ is the effect of one more treated positive-tie neighbor, averaged over the compositions an ego reaches at total $k$, with $R_k^{+}$ hypergeometric with population $N-1$, $a-1$ successes, and $k$ draws, and $\delta^{-}(t,k)=\mathbb E[\mu(t,R_k^{-},k-R_k^{-}+1)-\mu(t,R_k^{-},k-R_k^{-})]$ is its counterpart for one more treated negative-tie neighbor, with $R_k^{-}$ hypergeometric with population $N-1$, $a$ successes, and $k$ draws. No restriction is placed on $\mu$.
\end{proposition}

The two $\delta$ terms average tie-type increments over their respective composition distributions within the degree stratum. Their weighted sum can vanish when $a\delta^{+}(t,k)=-b\delta^{-}(t,k)$ even if both tie types have large effects. Thus a zero unsigned effect need not imply absence of peer influence.

\section{Estimation and Inference}
\label{sec:estimation}

SiDE predicts outcomes at the compared states and corrects them using observed prediction errors. Section~\ref{sec:score} weights these errors by inverse state probabilities, giving rarer states greater weight. Section~\ref{sec:encoder} fits the models and averages scores, and Section~\ref{sec:inference} estimates uncertainty from overlapping neighborhoods.

\subsection{Exposure propensities and doubly robust score}
\label{sec:score}

Each effect in Definitions~\ref{def:taupos}--\ref{def:kappa} is a \emph{contrast}: a weighted difference of mean potential outcomes. Write $z=(t,k^{+},k^{-})$, $Y_i(z)=Y_i(t,k^{+},k^{-})$, and use $m(z,q)$ from Eq.~\eqref{eq:m}. Effect $C$ has states $\mathcal Z_C$, coefficients $c_z$, and common reachable set $\mathcal V_C=\mathcal V(\max_z k^{+},\max_z k^{-})$ of size $n_C$:
\begin{equation}
\label{eq:contrast}
\tau_C=\mathbb E\Big[\sum_{z\in\mathcal Z_C}c_zY_i(z)\;\Big|\;i\in\mathcal V_C\Big].
\end{equation}
For the interaction in Definition~\ref{def:tauint}, the four coefficients are $+1,-1,-1,+1$ in the order of Eq.~\eqref{eq:tauint}. The score below applies the same coefficients to outcome predictions and residual corrections.

We model the binary treatments as mutually independent conditional on $(S,\mathbf X)$, with $p_j=\Pr(T_j=1\mid S,\mathbf X)\in(0,1)$. Since $\{i\}$, $\mathcal N_i^{+}$, and $\mathcal N_i^{-}$ are disjoint, this assignment model gives
\begin{equation}
\label{eq:pb}
e_i(z)=\Pr(Z_i=z\mid S,\mathbf X)=p_i^{\,t}(1-p_i)^{1-t}\;\mathrm{PB}\big(k^{+};\{p_j\}_{j\in\mathcal N_i^{+}}\big)\;\mathrm{PB}\big(k^{-};\{p_j\}_{j\in\mathcal N_i^{-}}\big),
\end{equation}
where $\mathrm{PB}(k;\{p_j\})$ is the Poisson--binomial probability that independent Bernoulli variables with success probabilities $\{p_j\}$ sum to $k$ \citep{hong2013}. Let $q_i$ be the observed value of $Q_i$. Conditional on $(S,\mathbf X)$, hold the candidate functions $(\bar m,\bar e)$ constant as treatments and outcomes vary. The AIPW score \citep{robins1994,cattaneo2010} is
\begin{equation}
\label{eq:score}
\varphi_C(O_i;\bar m,\bar e)
=\sum_{z\in\mathcal Z_C}c_z\left[\bar m(z,q_i)+\frac{\mathbf 1\{Z_i=z\}}{\bar e_i(z)}\big(Y_i-\bar m(z,q_i)\big)\right],
\end{equation}
where $O_i=(Y_i,Z_i,q_i)$. In Eq.~\eqref{eq:score}, the prediction error $Y_i-\bar m(z,q_i)$ is weighted by $\mathbf1\{Z_i=z\}/\bar e_i(z)$ to correct the outcome prediction.

\begin{proposition}[Double robustness]
\label{prop:dr}
Let Assumptions~\ref{as:exposure} and \ref{as:consistency} hold. Suppose $e_i(z)>0$ and $\mathbb E[Y_i(z)\mid Z_i=z,S,\mathbf X]=\mathbb E[Y_i(z)\mid S,\mathbf X]=m(z,q_i)$ for the states of $C$. Given $(S,\mathbf X)$, take $(\bar m,\bar e)$ to be nonrandom, with $\bar e_i(z)$ bounded away from zero. If $\bar m=m$ or $\bar e_i=e_i$, then $\mathbb E[\varphi_C(O_i;\bar m,\bar e)\mid S,\mathbf X]=\sum_{z}c_z\,m(z,q_i)$ for every ego of $\mathcal V_C$. The average score over $\mathcal V_C$ then has mean $\tau_C^{q}=n_C^{-1}\sum_{i\in\mathcal V_C}\sum_{z}c_z\,m(z,q_i)$ given the covariates, and the mean of $\tau_C^{q}$ over the covariates is $\tau_C$.
\end{proposition}

Taking the conditional expectation of the weighted residual in Eq.~\eqref{eq:score} establishes Proposition~\ref{prop:dr}.

\subsection{Sign-specific neighborhood encoder and estimator}
\label{sec:encoder}
\label{sec:estimator}

We implement SiDE's outcome and assignment models with separate encoders. Each uses pooling invariant to neighbor order \citep{zaheer2017} and distinguishes positive and negative neighborhoods, drawing on signed-network representation learning \citep{derr2018}:
\begin{equation}
\label{eq:encoder}
h_i=\rho\Big(X_i,\ \frac{1}{\max(d_i^{+},1)}\sum_{j\in\mathcal N_i^{+}}g^{+}(X_j),\ \frac{1}{\max(d_i^{-},1)}\sum_{j\in\mathcal N_i^{-}}g^{-}(X_j)\Big),
\end{equation}
where $g^{+}$, $g^{-}$, and $\rho$ are multilayer perceptrons. The outcome encoder gives $h_i$, and the assignment encoder has the same form with its own $g^{\pm}$ and $\rho$ and gives $\tilde h_i$. The outcome head combines $h_i$ with embeddings of the two exposure levels through main terms and a bilinear interaction. The assignment head uses a logistic function of $\tilde h_j$ to predict $p_j$, and Eq.~\eqref{eq:pb} turns these predictions into exposure propensities.

SiDE partitions target egos into $F$ folds and fits both models outside each fold before scoring its egos. With these held-out predictions, the unnormalized estimate is
\begin{equation}
\label{eq:aipw}
\check\tau_C=n_C^{-1}\sum_{i\in\mathcal V_C}\varphi_C(O_i;\hat m,\hat e).
\end{equation}
Normalizing the residual weights gives the single-partition estimate $\tilde\tau_C$, and SiDE reports the median $\hat\tau_C$ of $\tilde\tau_C$ over $P$ fold partitions,.

\subsection{Variance estimation and inference}
\label{sec:inference}

Define the closed neighborhood $\bar{\mathcal N}_i=\{i\}\cup\mathcal N_i$ of every ego, and call $i,j$ linked, written $i\sim j$, if $\bar{\mathcal N}_i\cap\bar{\mathcal N}_j\ne\varnothing$. Suppose that, given $(S,\mathbf X)$, unit-level disturbances are mutually independent and independent of treatments, and each potential outcome depends only on disturbances in its closed neighborhood. Under the conditional mean condition of Proposition~\ref{prop:dr}, the residuals using the true $m,e$,
\begin{equation}
\label{eq:resid}
r_i=\sum_{z\in\mathcal Z_C}c_z\frac{\mathbf1\{Z_i=z\}}{e_i(z)}\{Y_i-m(z,q_i)\},
\end{equation}
have mean zero and are independent for unlinked egos conditional on $(S,\mathbf X,\mathbf T)$. Replacing $m,e$ by their held-out estimates gives $\hat r_i$ and the linked-pair variance estimate
\begin{equation}
\label{eq:varhat}
\tilde\Omega_C=\frac{\sum_{i,j\in\mathcal V_C:\,i\sim j}\hat r_i\hat r_j}{n_C^2},
\end{equation}
where pairs include $i=j$ and $\hat r_i$ uses the same normalized inverse-probability weights as $\tilde\tau_C$.

\section{Experiments}
\label{sec:experiments}

In this section, we first evaluate SiDE on semi-synthetic data built on six real signed networks and compare it with eleven baselines. Then, we examine the role of tie signs, the sensitivity to the hidden width of the neural networks, and the computational cost, followed by an ablation study and an evaluation of interval coverage. Finally, we reanalyze data from a published school field experiment. Unless stated otherwise, each semi-synthetic experiment is repeated \mvRepMain{} times, and we report the average and the standard deviation.

\subsection{Experimental setup}
\label{sec:setup}

\begin{table}[t]
\centering
\caption{MAE on four networks for methods that use tie signs, with standard deviations of the absolute error below. Bold and underline mark the best and second-best values.}
\label{tab:main}
\scriptsize
\setlength{\tabcolsep}{1.2pt}
\begin{tabular}{lcccccccccccccccc}
\toprule
 & \multicolumn{4}{c}{Slashdot} & \multicolumn{4}{c}{Epinions} & \multicolumn{4}{c}{Reddit} & \multicolumn{4}{c}{Wiki-Elec} \\
Method & $\tau_{+}$ & $\tau_{-}$ & $\tau_{\mathrm{int}}$ & $\kappa$ & $\tau_{+}$ & $\tau_{-}$ & $\tau_{\mathrm{int}}$ & $\kappa$ & $\tau_{+}$ & $\tau_{-}$ & $\tau_{\mathrm{int}}$ & $\kappa$ & $\tau_{+}$ & $\tau_{-}$ & $\tau_{\mathrm{int}}$ & $\kappa$ \\
\midrule
Sat. reg. & \begin{tabular}[c]{@{}c@{}}0.059\\[-1.5pt]{\tiny$\pm$0.033}\end{tabular} & \begin{tabular}[c]{@{}c@{}}0.123\\[-1.5pt]{\tiny$\pm$0.048}\end{tabular} & \begin{tabular}[c]{@{}c@{}}0.068\\[-1.5pt]{\tiny$\pm$0.060}\end{tabular} & \begin{tabular}[c]{@{}c@{}}0.180\\[-1.5pt]{\tiny$\pm$0.042}\end{tabular} & \begin{tabular}[c]{@{}c@{}}0.056\\[-1.5pt]{\tiny$\pm$0.021}\end{tabular} & \begin{tabular}[c]{@{}c@{}}0.123\\[-1.5pt]{\tiny$\pm$0.048}\end{tabular} & \begin{tabular}[c]{@{}c@{}}0.044\\[-1.5pt]{\tiny$\pm$0.049}\end{tabular} & \begin{tabular}[c]{@{}c@{}}0.178\\[-1.5pt]{\tiny$\pm$0.055}\end{tabular} & \begin{tabular}[c]{@{}c@{}}0.048\\[-1.5pt]{\tiny$\pm$0.034}\end{tabular} & \begin{tabular}[c]{@{}c@{}}\underline{0.139}\\[-1.5pt]{\tiny$\pm$0.120}\end{tabular} & \begin{tabular}[c]{@{}c@{}}\underline{0.148}\\[-1.5pt]{\tiny$\pm$0.078}\end{tabular} & \begin{tabular}[c]{@{}c@{}}0.156\\[-1.5pt]{\tiny$\pm$0.108}\end{tabular} & \begin{tabular}[c]{@{}c@{}}0.064\\[-1.5pt]{\tiny$\pm$0.041}\end{tabular} & \begin{tabular}[c]{@{}c@{}}0.157\\[-1.5pt]{\tiny$\pm$0.089}\end{tabular} & \begin{tabular}[c]{@{}c@{}}0.143\\[-1.5pt]{\tiny$\pm$0.075}\end{tabular} & \begin{tabular}[c]{@{}c@{}}0.208\\[-1.5pt]{\tiny$\pm$0.128}\end{tabular} \\
GPS & \begin{tabular}[c]{@{}c@{}}0.100\\[-1.5pt]{\tiny$\pm$0.065}\end{tabular} & \begin{tabular}[c]{@{}c@{}}1.532\\[-1.5pt]{\tiny$\pm$0.074}\end{tabular} & \begin{tabular}[c]{@{}c@{}}0.270\\[-1.5pt]{\tiny$\pm$0.066}\end{tabular} & \begin{tabular}[c]{@{}c@{}}1.579\\[-1.5pt]{\tiny$\pm$0.077}\end{tabular} & \begin{tabular}[c]{@{}c@{}}0.109\\[-1.5pt]{\tiny$\pm$0.032}\end{tabular} & \begin{tabular}[c]{@{}c@{}}2.191\\[-1.5pt]{\tiny$\pm$0.065}\end{tabular} & \begin{tabular}[c]{@{}c@{}}0.405\\[-1.5pt]{\tiny$\pm$0.090}\end{tabular} & \begin{tabular}[c]{@{}c@{}}2.331\\[-1.5pt]{\tiny$\pm$0.064}\end{tabular} & \begin{tabular}[c]{@{}c@{}}0.880\\[-1.5pt]{\tiny$\pm$0.193}\end{tabular} & \begin{tabular}[c]{@{}c@{}}2.490\\[-1.5pt]{\tiny$\pm$0.243}\end{tabular} & \begin{tabular}[c]{@{}c@{}}0.384\\[-1.5pt]{\tiny$\pm$0.153}\end{tabular} & \begin{tabular}[c]{@{}c@{}}2.540\\[-1.5pt]{\tiny$\pm$0.297}\end{tabular} & \begin{tabular}[c]{@{}c@{}}0.154\\[-1.5pt]{\tiny$\pm$0.079}\end{tabular} & \begin{tabular}[c]{@{}c@{}}0.717\\[-1.5pt]{\tiny$\pm$0.167}\end{tabular} & \begin{tabular}[c]{@{}c@{}}0.235\\[-1.5pt]{\tiny$\pm$0.146}\end{tabular} & \begin{tabular}[c]{@{}c@{}}0.760\\[-1.5pt]{\tiny$\pm$0.154}\end{tabular} \\
TARNet & \begin{tabular}[c]{@{}c@{}}0.042\\[-1.5pt]{\tiny$\pm$0.018}\end{tabular} & \begin{tabular}[c]{@{}c@{}}\underline{0.045}\\[-1.5pt]{\tiny$\pm$0.029}\end{tabular} & \begin{tabular}[c]{@{}c@{}}\underline{0.048}\\[-1.5pt]{\tiny$\pm$0.029}\end{tabular} & \begin{tabular}[c]{@{}c@{}}\underline{0.099}\\[-1.5pt]{\tiny$\pm$0.030}\end{tabular} & \begin{tabular}[c]{@{}c@{}}0.019\\[-1.5pt]{\tiny$\pm$0.010}\end{tabular} & \begin{tabular}[c]{@{}c@{}}0.053\\[-1.5pt]{\tiny$\pm$0.017}\end{tabular} & \begin{tabular}[c]{@{}c@{}}0.170\\[-1.5pt]{\tiny$\pm$0.029}\end{tabular} & \begin{tabular}[c]{@{}c@{}}0.082\\[-1.5pt]{\tiny$\pm$0.031}\end{tabular} & \begin{tabular}[c]{@{}c@{}}\textbf{0.032}\\[-1.5pt]{\tiny$\pm$0.026}\end{tabular} & \begin{tabular}[c]{@{}c@{}}0.532\\[-1.5pt]{\tiny$\pm$0.083}\end{tabular} & \begin{tabular}[c]{@{}c@{}}0.298\\[-1.5pt]{\tiny$\pm$0.041}\end{tabular} & \begin{tabular}[c]{@{}c@{}}0.473\\[-1.5pt]{\tiny$\pm$0.081}\end{tabular} & \begin{tabular}[c]{@{}c@{}}0.046\\[-1.5pt]{\tiny$\pm$0.033}\end{tabular} & \begin{tabular}[c]{@{}c@{}}\underline{0.065}\\[-1.5pt]{\tiny$\pm$0.038}\end{tabular} & \begin{tabular}[c]{@{}c@{}}0.161\\[-1.5pt]{\tiny$\pm$0.023}\end{tabular} & \begin{tabular}[c]{@{}c@{}}0.191\\[-1.5pt]{\tiny$\pm$0.063}\end{tabular} \\
CFR & \begin{tabular}[c]{@{}c@{}}\underline{0.032}\\[-1.5pt]{\tiny$\pm$0.017}\end{tabular} & \begin{tabular}[c]{@{}c@{}}\textbf{0.036}\\[-1.5pt]{\tiny$\pm$0.024}\end{tabular} & \begin{tabular}[c]{@{}c@{}}\textbf{0.039}\\[-1.5pt]{\tiny$\pm$0.023}\end{tabular} & \begin{tabular}[c]{@{}c@{}}0.108\\[-1.5pt]{\tiny$\pm$0.038}\end{tabular} & \begin{tabular}[c]{@{}c@{}}0.017\\[-1.5pt]{\tiny$\pm$0.011}\end{tabular} & \begin{tabular}[c]{@{}c@{}}\underline{0.017}\\[-1.5pt]{\tiny$\pm$0.010}\end{tabular} & \begin{tabular}[c]{@{}c@{}}0.106\\[-1.5pt]{\tiny$\pm$0.028}\end{tabular} & \begin{tabular}[c]{@{}c@{}}\underline{0.075}\\[-1.5pt]{\tiny$\pm$0.024}\end{tabular} & \begin{tabular}[c]{@{}c@{}}0.036\\[-1.5pt]{\tiny$\pm$0.027}\end{tabular} & \begin{tabular}[c]{@{}c@{}}0.319\\[-1.5pt]{\tiny$\pm$0.072}\end{tabular} & \begin{tabular}[c]{@{}c@{}}0.240\\[-1.5pt]{\tiny$\pm$0.037}\end{tabular} & \begin{tabular}[c]{@{}c@{}}0.324\\[-1.5pt]{\tiny$\pm$0.066}\end{tabular} & \begin{tabular}[c]{@{}c@{}}0.043\\[-1.5pt]{\tiny$\pm$0.032}\end{tabular} & \begin{tabular}[c]{@{}c@{}}0.098\\[-1.5pt]{\tiny$\pm$0.064}\end{tabular} & \begin{tabular}[c]{@{}c@{}}0.176\\[-1.5pt]{\tiny$\pm$0.033}\end{tabular} & \begin{tabular}[c]{@{}c@{}}0.185\\[-1.5pt]{\tiny$\pm$0.088}\end{tabular} \\
SAGE-HSIC & \begin{tabular}[c]{@{}c@{}}0.134\\[-1.5pt]{\tiny$\pm$0.053}\end{tabular} & \begin{tabular}[c]{@{}c@{}}0.097\\[-1.5pt]{\tiny$\pm$0.042}\end{tabular} & \begin{tabular}[c]{@{}c@{}}0.162\\[-1.5pt]{\tiny$\pm$0.096}\end{tabular} & \begin{tabular}[c]{@{}c@{}}0.259\\[-1.5pt]{\tiny$\pm$0.057}\end{tabular} & \begin{tabular}[c]{@{}c@{}}0.073\\[-1.5pt]{\tiny$\pm$0.026}\end{tabular} & \begin{tabular}[c]{@{}c@{}}0.226\\[-1.5pt]{\tiny$\pm$0.047}\end{tabular} & \begin{tabular}[c]{@{}c@{}}0.096\\[-1.5pt]{\tiny$\pm$0.081}\end{tabular} & \begin{tabular}[c]{@{}c@{}}0.298\\[-1.5pt]{\tiny$\pm$0.069}\end{tabular} & \begin{tabular}[c]{@{}c@{}}0.440\\[-1.5pt]{\tiny$\pm$0.074}\end{tabular} & \begin{tabular}[c]{@{}c@{}}0.312\\[-1.5pt]{\tiny$\pm$0.279}\end{tabular} & \begin{tabular}[c]{@{}c@{}}0.245\\[-1.5pt]{\tiny$\pm$0.182}\end{tabular} & \begin{tabular}[c]{@{}c@{}}0.808\\[-1.5pt]{\tiny$\pm$0.312}\end{tabular} & \begin{tabular}[c]{@{}c@{}}0.131\\[-1.5pt]{\tiny$\pm$0.029}\end{tabular} & \begin{tabular}[c]{@{}c@{}}0.148\\[-1.5pt]{\tiny$\pm$0.090}\end{tabular} & \begin{tabular}[c]{@{}c@{}}\textbf{0.107}\\[-1.5pt]{\tiny$\pm$0.070}\end{tabular} & \begin{tabular}[c]{@{}c@{}}0.288\\[-1.5pt]{\tiny$\pm$0.099}\end{tabular} \\
NetEst & \begin{tabular}[c]{@{}c@{}}0.084\\[-1.5pt]{\tiny$\pm$0.043}\end{tabular} & \begin{tabular}[c]{@{}c@{}}0.444\\[-1.5pt]{\tiny$\pm$0.038}\end{tabular} & \begin{tabular}[c]{@{}c@{}}0.081\\[-1.5pt]{\tiny$\pm$0.048}\end{tabular} & \begin{tabular}[c]{@{}c@{}}0.709\\[-1.5pt]{\tiny$\pm$0.079}\end{tabular} & \begin{tabular}[c]{@{}c@{}}0.065\\[-1.5pt]{\tiny$\pm$0.019}\end{tabular} & \begin{tabular}[c]{@{}c@{}}0.383\\[-1.5pt]{\tiny$\pm$0.073}\end{tabular} & \begin{tabular}[c]{@{}c@{}}0.300\\[-1.5pt]{\tiny$\pm$0.044}\end{tabular} & \begin{tabular}[c]{@{}c@{}}0.614\\[-1.5pt]{\tiny$\pm$0.090}\end{tabular} & \begin{tabular}[c]{@{}c@{}}0.042\\[-1.5pt]{\tiny$\pm$0.029}\end{tabular} & \begin{tabular}[c]{@{}c@{}}0.788\\[-1.5pt]{\tiny$\pm$0.113}\end{tabular} & \begin{tabular}[c]{@{}c@{}}0.348\\[-1.5pt]{\tiny$\pm$0.035}\end{tabular} & \begin{tabular}[c]{@{}c@{}}0.835\\[-1.5pt]{\tiny$\pm$0.112}\end{tabular} & \begin{tabular}[c]{@{}c@{}}0.067\\[-1.5pt]{\tiny$\pm$0.035}\end{tabular} & \begin{tabular}[c]{@{}c@{}}0.329\\[-1.5pt]{\tiny$\pm$0.107}\end{tabular} & \begin{tabular}[c]{@{}c@{}}\underline{0.112}\\[-1.5pt]{\tiny$\pm$0.077}\end{tabular} & \begin{tabular}[c]{@{}c@{}}0.485\\[-1.5pt]{\tiny$\pm$0.112}\end{tabular} \\
HINITE & \begin{tabular}[c]{@{}c@{}}0.060\\[-1.5pt]{\tiny$\pm$0.028}\end{tabular} & \begin{tabular}[c]{@{}c@{}}0.628\\[-1.5pt]{\tiny$\pm$0.058}\end{tabular} & \begin{tabular}[c]{@{}c@{}}0.369\\[-1.5pt]{\tiny$\pm$0.048}\end{tabular} & \begin{tabular}[c]{@{}c@{}}0.559\\[-1.5pt]{\tiny$\pm$0.042}\end{tabular} & \begin{tabular}[c]{@{}c@{}}0.067\\[-1.5pt]{\tiny$\pm$0.015}\end{tabular} & \begin{tabular}[c]{@{}c@{}}0.470\\[-1.5pt]{\tiny$\pm$0.040}\end{tabular} & \begin{tabular}[c]{@{}c@{}}0.407\\[-1.5pt]{\tiny$\pm$0.029}\end{tabular} & \begin{tabular}[c]{@{}c@{}}0.451\\[-1.5pt]{\tiny$\pm$0.024}\end{tabular} & \begin{tabular}[c]{@{}c@{}}\underline{0.033}\\[-1.5pt]{\tiny$\pm$0.020}\end{tabular} & \begin{tabular}[c]{@{}c@{}}1.032\\[-1.5pt]{\tiny$\pm$0.085}\end{tabular} & \begin{tabular}[c]{@{}c@{}}0.415\\[-1.5pt]{\tiny$\pm$0.025}\end{tabular} & \begin{tabular}[c]{@{}c@{}}0.890\\[-1.5pt]{\tiny$\pm$0.077}\end{tabular} & \begin{tabular}[c]{@{}c@{}}\textbf{0.024}\\[-1.5pt]{\tiny$\pm$0.020}\end{tabular} & \begin{tabular}[c]{@{}c@{}}0.730\\[-1.5pt]{\tiny$\pm$0.142}\end{tabular} & \begin{tabular}[c]{@{}c@{}}0.390\\[-1.5pt]{\tiny$\pm$0.042}\end{tabular} & \begin{tabular}[c]{@{}c@{}}0.468\\[-1.5pt]{\tiny$\pm$0.116}\end{tabular} \\
DWR & \begin{tabular}[c]{@{}c@{}}0.056\\[-1.5pt]{\tiny$\pm$0.031}\end{tabular} & \begin{tabular}[c]{@{}c@{}}0.582\\[-1.5pt]{\tiny$\pm$0.062}\end{tabular} & \begin{tabular}[c]{@{}c@{}}0.230\\[-1.5pt]{\tiny$\pm$0.092}\end{tabular} & \begin{tabular}[c]{@{}c@{}}0.528\\[-1.5pt]{\tiny$\pm$0.088}\end{tabular} & \begin{tabular}[c]{@{}c@{}}0.028\\[-1.5pt]{\tiny$\pm$0.020}\end{tabular} & \begin{tabular}[c]{@{}c@{}}0.371\\[-1.5pt]{\tiny$\pm$0.029}\end{tabular} & \begin{tabular}[c]{@{}c@{}}0.423\\[-1.5pt]{\tiny$\pm$0.039}\end{tabular} & \begin{tabular}[c]{@{}c@{}}0.199\\[-1.5pt]{\tiny$\pm$0.059}\end{tabular} & \begin{tabular}[c]{@{}c@{}}0.049\\[-1.5pt]{\tiny$\pm$0.029}\end{tabular} & \begin{tabular}[c]{@{}c@{}}1.011\\[-1.5pt]{\tiny$\pm$0.118}\end{tabular} & \begin{tabular}[c]{@{}c@{}}0.389\\[-1.5pt]{\tiny$\pm$0.059}\end{tabular} & \begin{tabular}[c]{@{}c@{}}0.757\\[-1.5pt]{\tiny$\pm$0.150}\end{tabular} & \begin{tabular}[c]{@{}c@{}}\underline{0.039}\\[-1.5pt]{\tiny$\pm$0.027}\end{tabular} & \begin{tabular}[c]{@{}c@{}}0.462\\[-1.5pt]{\tiny$\pm$0.128}\end{tabular} & \begin{tabular}[c]{@{}c@{}}0.125\\[-1.5pt]{\tiny$\pm$0.094}\end{tabular} & \begin{tabular}[c]{@{}c@{}}0.301\\[-1.5pt]{\tiny$\pm$0.126}\end{tabular} \\
GNN-DR & \begin{tabular}[c]{@{}c@{}}0.043\\[-1.5pt]{\tiny$\pm$0.031}\end{tabular} & \begin{tabular}[c]{@{}c@{}}0.328\\[-1.5pt]{\tiny$\pm$0.207}\end{tabular} & \begin{tabular}[c]{@{}c@{}}0.330\\[-1.5pt]{\tiny$\pm$0.195}\end{tabular} & \begin{tabular}[c]{@{}c@{}}0.480\\[-1.5pt]{\tiny$\pm$0.282}\end{tabular} & \begin{tabular}[c]{@{}c@{}}0.030\\[-1.5pt]{\tiny$\pm$0.020}\end{tabular} & \begin{tabular}[c]{@{}c@{}}0.156\\[-1.5pt]{\tiny$\pm$0.082}\end{tabular} & \begin{tabular}[c]{@{}c@{}}0.220\\[-1.5pt]{\tiny$\pm$0.133}\end{tabular} & \begin{tabular}[c]{@{}c@{}}0.413\\[-1.5pt]{\tiny$\pm$0.174}\end{tabular} & \begin{tabular}[c]{@{}c@{}}0.126\\[-1.5pt]{\tiny$\pm$0.128}\end{tabular} & \begin{tabular}[c]{@{}c@{}}1.003\\[-1.5pt]{\tiny$\pm$0.764}\end{tabular} & \begin{tabular}[c]{@{}c@{}}0.938\\[-1.5pt]{\tiny$\pm$0.701}\end{tabular} & \begin{tabular}[c]{@{}c@{}}1.966\\[-1.5pt]{\tiny$\pm$1.224}\end{tabular} & \begin{tabular}[c]{@{}c@{}}0.099\\[-1.5pt]{\tiny$\pm$0.064}\end{tabular} & \begin{tabular}[c]{@{}c@{}}0.322\\[-1.5pt]{\tiny$\pm$0.283}\end{tabular} & \begin{tabular}[c]{@{}c@{}}0.586\\[-1.5pt]{\tiny$\pm$0.746}\end{tabular} & \begin{tabular}[c]{@{}c@{}}0.493\\[-1.5pt]{\tiny$\pm$0.410}\end{tabular} \\
HINet & \begin{tabular}[c]{@{}c@{}}0.037\\[-1.5pt]{\tiny$\pm$0.051}\end{tabular} & \begin{tabular}[c]{@{}c@{}}0.313\\[-1.5pt]{\tiny$\pm$0.062}\end{tabular} & \begin{tabular}[c]{@{}c@{}}0.295\\[-1.5pt]{\tiny$\pm$0.036}\end{tabular} & \begin{tabular}[c]{@{}c@{}}0.150\\[-1.5pt]{\tiny$\pm$0.073}\end{tabular} & \begin{tabular}[c]{@{}c@{}}\underline{0.017}\\[-1.5pt]{\tiny$\pm$0.012}\end{tabular} & \begin{tabular}[c]{@{}c@{}}0.320\\[-1.5pt]{\tiny$\pm$0.091}\end{tabular} & \begin{tabular}[c]{@{}c@{}}0.346\\[-1.5pt]{\tiny$\pm$0.041}\end{tabular} & \begin{tabular}[c]{@{}c@{}}0.167\\[-1.5pt]{\tiny$\pm$0.046}\end{tabular} & \begin{tabular}[c]{@{}c@{}}0.067\\[-1.5pt]{\tiny$\pm$0.036}\end{tabular} & \begin{tabular}[c]{@{}c@{}}0.956\\[-1.5pt]{\tiny$\pm$0.133}\end{tabular} & \begin{tabular}[c]{@{}c@{}}0.425\\[-1.5pt]{\tiny$\pm$0.017}\end{tabular} & \begin{tabular}[c]{@{}c@{}}0.662\\[-1.5pt]{\tiny$\pm$0.127}\end{tabular} & \begin{tabular}[c]{@{}c@{}}0.055\\[-1.5pt]{\tiny$\pm$0.044}\end{tabular} & \begin{tabular}[c]{@{}c@{}}0.250\\[-1.5pt]{\tiny$\pm$0.116}\end{tabular} & \begin{tabular}[c]{@{}c@{}}0.280\\[-1.5pt]{\tiny$\pm$0.039}\end{tabular} & \begin{tabular}[c]{@{}c@{}}\underline{0.103}\\[-1.5pt]{\tiny$\pm$0.047}\end{tabular} \\
GAIPW & \begin{tabular}[c]{@{}c@{}}0.067\\[-1.5pt]{\tiny$\pm$0.038}\end{tabular} & \begin{tabular}[c]{@{}c@{}}0.139\\[-1.5pt]{\tiny$\pm$0.065}\end{tabular} & \begin{tabular}[c]{@{}c@{}}0.116\\[-1.5pt]{\tiny$\pm$0.085}\end{tabular} & \begin{tabular}[c]{@{}c@{}}0.204\\[-1.5pt]{\tiny$\pm$0.090}\end{tabular} & \begin{tabular}[c]{@{}c@{}}0.060\\[-1.5pt]{\tiny$\pm$0.022}\end{tabular} & \begin{tabular}[c]{@{}c@{}}0.157\\[-1.5pt]{\tiny$\pm$0.048}\end{tabular} & \begin{tabular}[c]{@{}c@{}}\underline{0.039}\\[-1.5pt]{\tiny$\pm$0.024}\end{tabular} & \begin{tabular}[c]{@{}c@{}}0.215\\[-1.5pt]{\tiny$\pm$0.053}\end{tabular} & \begin{tabular}[c]{@{}c@{}}0.042\\[-1.5pt]{\tiny$\pm$0.040}\end{tabular} & \begin{tabular}[c]{@{}c@{}}0.155\\[-1.5pt]{\tiny$\pm$0.064}\end{tabular} & \begin{tabular}[c]{@{}c@{}}0.191\\[-1.5pt]{\tiny$\pm$0.169}\end{tabular} & \begin{tabular}[c]{@{}c@{}}\underline{0.152}\\[-1.5pt]{\tiny$\pm$0.107}\end{tabular} & \begin{tabular}[c]{@{}c@{}}0.064\\[-1.5pt]{\tiny$\pm$0.047}\end{tabular} & \begin{tabular}[c]{@{}c@{}}0.145\\[-1.5pt]{\tiny$\pm$0.116}\end{tabular} & \begin{tabular}[c]{@{}c@{}}0.361\\[-1.5pt]{\tiny$\pm$0.301}\end{tabular} & \begin{tabular}[c]{@{}c@{}}0.149\\[-1.5pt]{\tiny$\pm$0.117}\end{tabular} \\
\midrule
SiDE (ours) & \begin{tabular}[c]{@{}c@{}}\textbf{0.013}\\[-1.5pt]{\tiny$\pm$0.009}\end{tabular} & \begin{tabular}[c]{@{}c@{}}0.049\\[-1.5pt]{\tiny$\pm$0.028}\end{tabular} & \begin{tabular}[c]{@{}c@{}}0.085\\[-1.5pt]{\tiny$\pm$0.045}\end{tabular} & \begin{tabular}[c]{@{}c@{}}\textbf{0.064}\\[-1.5pt]{\tiny$\pm$0.034}\end{tabular} & \begin{tabular}[c]{@{}c@{}}\textbf{0.012}\\[-1.5pt]{\tiny$\pm$0.008}\end{tabular} & \begin{tabular}[c]{@{}c@{}}\textbf{0.014}\\[-1.5pt]{\tiny$\pm$0.009}\end{tabular} & \begin{tabular}[c]{@{}c@{}}\textbf{0.032}\\[-1.5pt]{\tiny$\pm$0.033}\end{tabular} & \begin{tabular}[c]{@{}c@{}}\textbf{0.027}\\[-1.5pt]{\tiny$\pm$0.018}\end{tabular} & \begin{tabular}[c]{@{}c@{}}0.042\\[-1.5pt]{\tiny$\pm$0.031}\end{tabular} & \begin{tabular}[c]{@{}c@{}}\textbf{0.118}\\[-1.5pt]{\tiny$\pm$0.075}\end{tabular} & \begin{tabular}[c]{@{}c@{}}\textbf{0.134}\\[-1.5pt]{\tiny$\pm$0.104}\end{tabular} & \begin{tabular}[c]{@{}c@{}}\textbf{0.128}\\[-1.5pt]{\tiny$\pm$0.083}\end{tabular} & \begin{tabular}[c]{@{}c@{}}0.042\\[-1.5pt]{\tiny$\pm$0.022}\end{tabular} & \begin{tabular}[c]{@{}c@{}}\textbf{0.060}\\[-1.5pt]{\tiny$\pm$0.052}\end{tabular} & \begin{tabular}[c]{@{}c@{}}0.195\\[-1.5pt]{\tiny$\pm$0.135}\end{tabular} & \begin{tabular}[c]{@{}c@{}}\textbf{0.093}\\[-1.5pt]{\tiny$\pm$0.097}\end{tabular} \\
\bottomrule
\end{tabular}

\end{table}

\textbf{Datasets and settings.} We use six real signed networks: Bitcoin-Alpha, Bitcoin-OTC, Wiki-Elec, Slashdot, Epinions, and Reddit. The two Bitcoin networks record ratings between traders \citep{kumar2016}. Wiki-Elec, Slashdot, and Epinions record votes, friend/foe relations, and trust/distrust, respectively \citep{leskovec2010}, and Reddit records signed hyperlinks between communities \citep{kumar2018community}. Following prior evaluations of network causal estimators \citep{ma2021}, we keep each observed network and generate treatments and outcomes on it, so that the true effects are known. For the first five networks, the six base covariates are standard normal draws independent of the network. For Reddit, they are the first six principal components of the released community embeddings. All models also receive signed degrees. In the two Bitcoin networks, few egos reach the exposure states compared by the interaction, so the main comparison uses the other four networks.

\textbf{Baselines.} We compare SiDE with eleven baselines from four families. Saturated regression~\citep{vazquezbare2023} and GPS~\citep{forastiere2021} adjust for exposure counts through regression and generalized propensity scores. TARNet and CFR~\citep{shalit2017} are neural treatment-effect estimators that take the exposures as inputs. SAGE-HSIC~\citep{ma2021}, NetEst~\citep{jiang2022}, HINITE~\citep{lin2023}, DWR~\citep{zhao2024}, GNN-DR~\citep{leung2025}, and HINet~\citep{caljon2026} are graph neural estimators for network interference, and GAIPW~\citep{qu2024} is a generalized AIPW estimator for type-specific exposures.

\textbf{Metrics.} We evaluate the estimates by the bias $J^{-1}\sum_{s=1}^{J}(\hat\tau_C^{(s)}-\tau_C^q)$ and the MAE (mean absolute error) $J^{-1}\sum_{s=1}^{J}|\hat\tau_C^{(s)}-\tau_C^q|$ over $J=\mvRepMain$ replications, where $\hat\tau_C^{(s)}$ is the estimate of contrast $C$ in replication $s$ and $\tau_C^q$ its true value. Lower MAE and smaller absolute bias indicate better estimation accuracy.

\subsection{Results on semi-synthetic data}
\label{sec:results}

\textbf{Effect estimation.} In Table~\ref{tab:main}, SiDE has the lowest MAE averaged over the four networks and four effects. It has the lowest MAE for all four effects on Epinions, and individual baselines are more accurate for some effects on the other three networks, including the interaction on Wiki-Elec. Across all six networks and four effects, SiDE also ranks first in average absolute bias.

\begin{figure}[t]
\centering
\includegraphics[width=\textwidth]{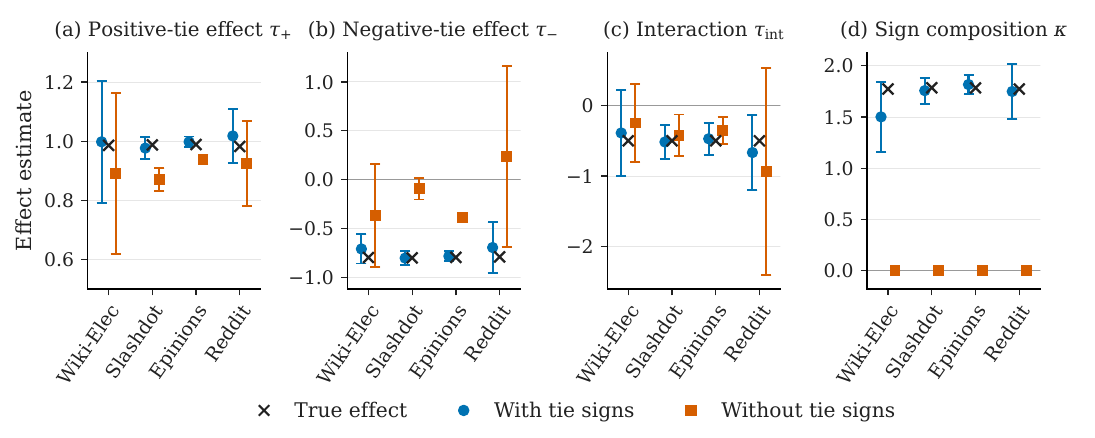}
\caption{Effects with and without tie signs. Markers show mean estimates, bars show standard deviations, and crosses show true effects.}
\label{fig:signed_effects}
\end{figure}

\textbf{Why tie signs matter.} As shown in Figure~\ref{fig:signed_effects}, removing the signs from an otherwise identical estimator attenuates the estimated negative-tie effect on Wiki-Elec, Slashdot, and Epinions and reverses its sign on average on Reddit. This doubly robust estimator combines SiDE's outcome network with a softmax exposure-state model, and its unsigned version, fitted on the same folds, pools positive and negative neighbors in both models. The unsigned version also makes the sign-composition effect identically zero, since the two composition states have the same treated-neighbor total.

\begin{table}[!t]
\centering
\caption{Ablation results on four networks: (a) MAE with standard deviations of the absolute error below and (b) bias. The variant w/o weighting drops the residual correction, w/o outcome drops the outcome model, and w/o PB predicts joint-state probabilities with a softmax model. Bold and underline mark the smallest and second-smallest MAE in (a) and absolute bias in (b).}
\label{tab:ablation_main}
\scriptsize
\setlength{\tabcolsep}{1.0pt}
\begin{tabular}{lcccccccccccccccc}
\toprule
\multicolumn{17}{l}{\textit{(a) MAE and standard deviation}} \\
 & \multicolumn{4}{c}{Slashdot} & \multicolumn{4}{c}{Epinions} & \multicolumn{4}{c}{Reddit} & \multicolumn{4}{c}{Wiki-Elec} \\
Variant & $\tau_{+}$ & $\tau_{-}$ & $\tau_{\mathrm{int}}$ & $\kappa$ & $\tau_{+}$ & $\tau_{-}$ & $\tau_{\mathrm{int}}$ & $\kappa$ & $\tau_{+}$ & $\tau_{-}$ & $\tau_{\mathrm{int}}$ & $\kappa$ & $\tau_{+}$ & $\tau_{-}$ & $\tau_{\mathrm{int}}$ & $\kappa$ \\
\midrule
SiDE & \begin{tabular}[c]{@{}c@{}}\textbf{0.013}\\[-1.5pt]{\tiny$\pm$0.009}\end{tabular} & \begin{tabular}[c]{@{}c@{}}\underline{0.049}\\[-1.5pt]{\tiny$\pm$0.028}\end{tabular} & \begin{tabular}[c]{@{}c@{}}\underline{0.085}\\[-1.5pt]{\tiny$\pm$0.045}\end{tabular} & \begin{tabular}[c]{@{}c@{}}\underline{0.064}\\[-1.5pt]{\tiny$\pm$0.034}\end{tabular} & \begin{tabular}[c]{@{}c@{}}\textbf{0.012}\\[-1.5pt]{\tiny$\pm$0.008}\end{tabular} & \begin{tabular}[c]{@{}c@{}}\textbf{0.014}\\[-1.5pt]{\tiny$\pm$0.009}\end{tabular} & \begin{tabular}[c]{@{}c@{}}\textbf{0.032}\\[-1.5pt]{\tiny$\pm$0.033}\end{tabular} & \begin{tabular}[c]{@{}c@{}}\textbf{0.027}\\[-1.5pt]{\tiny$\pm$0.018}\end{tabular} & \begin{tabular}[c]{@{}c@{}}\underline{0.042}\\[-1.5pt]{\tiny$\pm$0.031}\end{tabular} & \begin{tabular}[c]{@{}c@{}}\underline{0.118}\\[-1.5pt]{\tiny$\pm$0.075}\end{tabular} & \begin{tabular}[c]{@{}c@{}}\underline{0.134}\\[-1.5pt]{\tiny$\pm$0.104}\end{tabular} & \begin{tabular}[c]{@{}c@{}}\textbf{0.128}\\[-1.5pt]{\tiny$\pm$0.083}\end{tabular} & \begin{tabular}[c]{@{}c@{}}\underline{0.042}\\[-1.5pt]{\tiny$\pm$0.022}\end{tabular} & \begin{tabular}[c]{@{}c@{}}\textbf{0.060}\\[-1.5pt]{\tiny$\pm$0.052}\end{tabular} & \begin{tabular}[c]{@{}c@{}}\underline{0.195}\\[-1.5pt]{\tiny$\pm$0.135}\end{tabular} & \begin{tabular}[c]{@{}c@{}}\textbf{0.093}\\[-1.5pt]{\tiny$\pm$0.097}\end{tabular} \\
w/o weighting & \begin{tabular}[c]{@{}c@{}}\underline{0.015}\\[-1.5pt]{\tiny$\pm$0.004}\end{tabular} & \begin{tabular}[c]{@{}c@{}}\textbf{0.035}\\[-1.5pt]{\tiny$\pm$0.021}\end{tabular} & \begin{tabular}[c]{@{}c@{}}\textbf{0.053}\\[-1.5pt]{\tiny$\pm$0.046}\end{tabular} & \begin{tabular}[c]{@{}c@{}}\textbf{0.050}\\[-1.5pt]{\tiny$\pm$0.023}\end{tabular} & \begin{tabular}[c]{@{}c@{}}\underline{0.012}\\[-1.5pt]{\tiny$\pm$0.009}\end{tabular} & \begin{tabular}[c]{@{}c@{}}\underline{0.017}\\[-1.5pt]{\tiny$\pm$0.011}\end{tabular} & \begin{tabular}[c]{@{}c@{}}\underline{0.060}\\[-1.5pt]{\tiny$\pm$0.038}\end{tabular} & \begin{tabular}[c]{@{}c@{}}\underline{0.035}\\[-1.5pt]{\tiny$\pm$0.022}\end{tabular} & \begin{tabular}[c]{@{}c@{}}\textbf{0.040}\\[-1.5pt]{\tiny$\pm$0.027}\end{tabular} & \begin{tabular}[c]{@{}c@{}}\textbf{0.073}\\[-1.5pt]{\tiny$\pm$0.055}\end{tabular} & \begin{tabular}[c]{@{}c@{}}\textbf{0.112}\\[-1.5pt]{\tiny$\pm$0.089}\end{tabular} & \begin{tabular}[c]{@{}c@{}}\underline{0.143}\\[-1.5pt]{\tiny$\pm$0.060}\end{tabular} & \begin{tabular}[c]{@{}c@{}}\textbf{0.033}\\[-1.5pt]{\tiny$\pm$0.025}\end{tabular} & \begin{tabular}[c]{@{}c@{}}\underline{0.086}\\[-1.5pt]{\tiny$\pm$0.074}\end{tabular} & \begin{tabular}[c]{@{}c@{}}\textbf{0.121}\\[-1.5pt]{\tiny$\pm$0.065}\end{tabular} & \begin{tabular}[c]{@{}c@{}}\underline{0.167}\\[-1.5pt]{\tiny$\pm$0.117}\end{tabular} \\
w/o outcome & \begin{tabular}[c]{@{}c@{}}0.114\\[-1.5pt]{\tiny$\pm$0.109}\end{tabular} & \begin{tabular}[c]{@{}c@{}}0.156\\[-1.5pt]{\tiny$\pm$0.145}\end{tabular} & \begin{tabular}[c]{@{}c@{}}0.218\\[-1.5pt]{\tiny$\pm$0.165}\end{tabular} & \begin{tabular}[c]{@{}c@{}}0.248\\[-1.5pt]{\tiny$\pm$0.293}\end{tabular} & \begin{tabular}[c]{@{}c@{}}0.060\\[-1.5pt]{\tiny$\pm$0.028}\end{tabular} & \begin{tabular}[c]{@{}c@{}}0.147\\[-1.5pt]{\tiny$\pm$0.102}\end{tabular} & \begin{tabular}[c]{@{}c@{}}0.233\\[-1.5pt]{\tiny$\pm$0.273}\end{tabular} & \begin{tabular}[c]{@{}c@{}}0.156\\[-1.5pt]{\tiny$\pm$0.116}\end{tabular} & \begin{tabular}[c]{@{}c@{}}0.238\\[-1.5pt]{\tiny$\pm$0.108}\end{tabular} & \begin{tabular}[c]{@{}c@{}}0.829\\[-1.5pt]{\tiny$\pm$0.674}\end{tabular} & \begin{tabular}[c]{@{}c@{}}0.507\\[-1.5pt]{\tiny$\pm$0.436}\end{tabular} & \begin{tabular}[c]{@{}c@{}}0.991\\[-1.5pt]{\tiny$\pm$0.557}\end{tabular} & \begin{tabular}[c]{@{}c@{}}0.115\\[-1.5pt]{\tiny$\pm$0.053}\end{tabular} & \begin{tabular}[c]{@{}c@{}}0.243\\[-1.5pt]{\tiny$\pm$0.204}\end{tabular} & \begin{tabular}[c]{@{}c@{}}0.369\\[-1.5pt]{\tiny$\pm$0.400}\end{tabular} & \begin{tabular}[c]{@{}c@{}}0.319\\[-1.5pt]{\tiny$\pm$0.194}\end{tabular} \\
w/o PB & \begin{tabular}[c]{@{}c@{}}0.025\\[-1.5pt]{\tiny$\pm$0.024}\end{tabular} & \begin{tabular}[c]{@{}c@{}}0.138\\[-1.5pt]{\tiny$\pm$0.103}\end{tabular} & \begin{tabular}[c]{@{}c@{}}0.204\\[-1.5pt]{\tiny$\pm$0.132}\end{tabular} & \begin{tabular}[c]{@{}c@{}}0.106\\[-1.5pt]{\tiny$\pm$0.075}\end{tabular} & \begin{tabular}[c]{@{}c@{}}0.014\\[-1.5pt]{\tiny$\pm$0.009}\end{tabular} & \begin{tabular}[c]{@{}c@{}}0.042\\[-1.5pt]{\tiny$\pm$0.023}\end{tabular} & \begin{tabular}[c]{@{}c@{}}0.186\\[-1.5pt]{\tiny$\pm$0.080}\end{tabular} & \begin{tabular}[c]{@{}c@{}}0.091\\[-1.5pt]{\tiny$\pm$0.066}\end{tabular} & \begin{tabular}[c]{@{}c@{}}0.189\\[-1.5pt]{\tiny$\pm$0.161}\end{tabular} & \begin{tabular}[c]{@{}c@{}}0.714\\[-1.5pt]{\tiny$\pm$0.858}\end{tabular} & \begin{tabular}[c]{@{}c@{}}0.842\\[-1.5pt]{\tiny$\pm$0.890}\end{tabular} & \begin{tabular}[c]{@{}c@{}}0.378\\[-1.5pt]{\tiny$\pm$0.397}\end{tabular} & \begin{tabular}[c]{@{}c@{}}0.131\\[-1.5pt]{\tiny$\pm$0.294}\end{tabular} & \begin{tabular}[c]{@{}c@{}}0.164\\[-1.5pt]{\tiny$\pm$0.120}\end{tabular} & \begin{tabular}[c]{@{}c@{}}0.436\\[-1.5pt]{\tiny$\pm$0.302}\end{tabular} & \begin{tabular}[c]{@{}c@{}}0.281\\[-1.5pt]{\tiny$\pm$0.112}\end{tabular} \\
\midrule
\addlinespace[2pt]
\multicolumn{17}{l}{\textit{(b) Bias}} \\
 & \multicolumn{4}{c}{Slashdot} & \multicolumn{4}{c}{Epinions} & \multicolumn{4}{c}{Reddit} & \multicolumn{4}{c}{Wiki-Elec} \\
Variant & $\tau_{+}$ & $\tau_{-}$ & $\tau_{\mathrm{int}}$ & $\kappa$ & $\tau_{+}$ & $\tau_{-}$ & $\tau_{\mathrm{int}}$ & $\kappa$ & $\tau_{+}$ & $\tau_{-}$ & $\tau_{\mathrm{int}}$ & $\kappa$ & $\tau_{+}$ & $\tau_{-}$ & $\tau_{\mathrm{int}}$ & $\kappa$ \\
\midrule
SiDE & {\fontsize{6}{7}\selectfont \underline{$+$0.003}} & {\fontsize{6}{7}\selectfont {\boldmath$-$}\textbf{0.001}} & {\fontsize{6}{7}\selectfont {\boldmath$-$}\textbf{0.001}} & {\fontsize{6}{7}\selectfont {\boldmath$-$}\textbf{0.002}} & {\fontsize{6}{7}\selectfont {\boldmath$+$}\textbf{0.006}} & {\fontsize{6}{7}\selectfont \underline{$+$0.009}} & {\fontsize{6}{7}\selectfont {\boldmath$+$}\textbf{0.021}} & {\fontsize{6}{7}\selectfont {\boldmath$+$}\textbf{0.003}} & {\fontsize{6}{7}\selectfont {\boldmath$-$}\textbf{0.017}} & {\fontsize{6}{7}\selectfont {\boldmath$+$}\textbf{0.001}} & {\fontsize{6}{7}\selectfont \underline{$-$0.031}} & {\fontsize{6}{7}\selectfont {\boldmath$-$}\textbf{0.051}} & {\fontsize{6}{7}\selectfont {\boldmath$+$}\textbf{0.015}} & {\fontsize{6}{7}\selectfont {\boldmath$+$}\textbf{0.055}} & {\fontsize{6}{7}\selectfont \underline{$-$0.063}} & {\fontsize{6}{7}\selectfont \underline{$-$0.050}} \\
w/o weighting & {\fontsize{6}{7}\selectfont $+$0.004} & {\fontsize{6}{7}\selectfont $+$0.026} & {\fontsize{6}{7}\selectfont $+$0.045} & {\fontsize{6}{7}\selectfont \underline{$-$0.013}} & {\fontsize{6}{7}\selectfont \underline{$+$0.008}} & {\fontsize{6}{7}\selectfont $+$0.013} & {\fontsize{6}{7}\selectfont \underline{$+$0.058}} & {\fontsize{6}{7}\selectfont $-$0.015} & {\fontsize{6}{7}\selectfont \underline{$-$0.026}} & {\fontsize{6}{7}\selectfont \underline{$+$0.021}} & {\fontsize{6}{7}\selectfont $+$0.105} & {\fontsize{6}{7}\selectfont $-$0.143} & {\fontsize{6}{7}\selectfont \underline{$-$0.033}} & {\fontsize{6}{7}\selectfont \underline{$+$0.073}} & {\fontsize{6}{7}\selectfont $+$0.112} & {\fontsize{6}{7}\selectfont $-$0.167} \\
w/o outcome & {\fontsize{6}{7}\selectfont {\boldmath$+$}\textbf{0.002}} & {\fontsize{6}{7}\selectfont \underline{$-$0.007}} & {\fontsize{6}{7}\selectfont \underline{$-$0.026}} & {\fontsize{6}{7}\selectfont $+$0.025} & {\fontsize{6}{7}\selectfont $+$0.034} & {\fontsize{6}{7}\selectfont $-$0.040} & {\fontsize{6}{7}\selectfont $+$0.172} & {\fontsize{6}{7}\selectfont \underline{$+$0.009}} & {\fontsize{6}{7}\selectfont $+$0.159} & {\fontsize{6}{7}\selectfont $+$0.261} & {\fontsize{6}{7}\selectfont {\boldmath$-$}\textbf{0.012}} & {\fontsize{6}{7}\selectfont \underline{$+$0.099}} & {\fontsize{6}{7}\selectfont $+$0.070} & {\fontsize{6}{7}\selectfont $-$0.121} & {\fontsize{6}{7}\selectfont {\boldmath$-$}\textbf{0.021}} & {\fontsize{6}{7}\selectfont {\boldmath$+$}\textbf{0.026}} \\
w/o PB & {\fontsize{6}{7}\selectfont $-$0.014} & {\fontsize{6}{7}\selectfont $-$0.088} & {\fontsize{6}{7}\selectfont $+$0.083} & {\fontsize{6}{7}\selectfont $-$0.015} & {\fontsize{6}{7}\selectfont $+$0.013} & {\fontsize{6}{7}\selectfont {\boldmath$+$}\textbf{0.006}} & {\fontsize{6}{7}\selectfont $+$0.077} & {\fontsize{6}{7}\selectfont $+$0.022} & {\fontsize{6}{7}\selectfont $-$0.044} & {\fontsize{6}{7}\selectfont $+$0.163} & {\fontsize{6}{7}\selectfont $+$0.106} & {\fontsize{6}{7}\selectfont $-$0.198} & {\fontsize{6}{7}\selectfont $+$0.078} & {\fontsize{6}{7}\selectfont $+$0.148} & {\fontsize{6}{7}\selectfont $-$0.256} & {\fontsize{6}{7}\selectfont $-$0.281} \\
\bottomrule
\end{tabular}

\end{table}

\textbf{Sensitivity to hidden width.} SiDE keeps the lowest average MAE among the methods in Table~\ref{tab:main} when the hidden width of the outcome and assignment models, the number of units in each hidden layer, is 32, 64, or 128. On the four main networks, the MAE of each effect changes by less than two Monte Carlo standard errors across these widths. The accuracy of SiDE is therefore stable across this range of model capacity.

\textbf{Computational cost.} With the depth of each multilayer perceptron fixed, each training epoch costs $O(|E|H+nH^2)$ for $n$ units, the tie set $E=\{(i,j):S_{ij}\ne0\}$, and hidden width $H$. One complete SiDE run takes \timeFast--\timeSlow\ seconds after preprocessing on one NVIDIA RTX 6000 Ada GPU.

\par
\subsection{Ablation study}
\label{sec:ablation}

In Table~\ref{tab:ablation_main}, dropping the outcome model (w/o outcome) or the Poisson-binomial exposure probabilities (w/o PB) raises the MAE for every network and effect. Dropping the residual correction (w/o weighting) gives a similar MAE but a larger bias, with average absolute bias \mvAblMeanBiasOutcome{} against \mvAblMeanBiasSide{} for SiDE. SiDE's bias stays within two Monte Carlo standard errors of zero for every network and effect except the negative-tie effect on Wiki-Elec. The price is a larger standard deviation of the estimates, \mvAblMeanSdSide{} against \mvAblMeanSdOutcome.

\begin{table}[t]
\centering
\caption{Coverage over \mvRepCoverage{} replications with 95\% Clopper--Pearson (CP) confidence intervals. Underlining marks coverage below \mvCovThreshold, two binomial standard errors below 0.95.}
\label{tab:coverage}
\scriptsize
\setlength{\tabcolsep}{4pt}
\begin{tabular}{lcccccccc}
\toprule
Network & \multicolumn{2}{c}{$\tau_{+}$} & \multicolumn{2}{c}{$\tau_{-}$} & \multicolumn{2}{c}{$\tau_{\mathrm{int}}$} & \multicolumn{2}{c}{$\kappa$} \\
\cmidrule(lr){2-3}\cmidrule(lr){4-5}\cmidrule(lr){6-7}\cmidrule(lr){8-9}
 & Coverage & 95\% CP interval & Coverage & 95\% CP interval & Coverage & 95\% CP interval & Coverage & 95\% CP interval \\
\midrule
Slashdot & 0.930 & [0.885, 0.961] & 0.955 & [0.916, 0.979] & 0.965 & [0.929, 0.986] & 0.945 & [0.904, 0.972] \\
Epinions & \underline{0.905} & [0.856, 0.942] & 0.930 & [0.885, 0.961] & 0.965 & [0.929, 0.986] & 0.965 & [0.929, 0.986] \\
Reddit & \underline{0.850} & [0.793, 0.896] & 0.935 & [0.891, 0.965] & 0.945 & [0.904, 0.972] & 0.985 & [0.957, 0.997] \\
Wiki-Elec & 0.960 & [0.923, 0.983] & 0.920 & [0.873, 0.954] & 0.970 & [0.936, 0.989] & 0.940 & [0.898, 0.969] \\
\bottomrule
\end{tabular}

\end{table}

\needlines{6\baselineskip}
\subsection{Coverage of approximate intervals}
\label{sec:coverage}

Table~\ref{tab:coverage} reports coverage of at least \mvCovThreshold, two binomial standard errors below 0.95, for every network and effect except the positive-tie effect on Epinions and Reddit. We generate \mvRepCoverage{} datasets on each of the four main networks, with independent treatment assignment and disturbances shared through overlapping neighborhoods, and coverage is the fraction of datasets whose nominal 95\% interval $\hat\tau_C\pm1.96\sqrt{\hat\Omega_C}$, with $\hat\Omega_C$ aggregated over partitions, contains the known effect. On Reddit, the positive-tie coverage is \vcMinCov, and on the two Bitcoin networks the positive-tie intervals also undercover, with coverage \vbPosAlphaCov{} and \vbPosOtcCov.

\subsection{Case study}
\label{sec:case}

In the \caseMainSchoolCount{} program schools of the anti-conflict experiment of \citet{paluck2016,paluck2020data}, a student is treated if randomly selected for an invitation to the program, whether or not the student participated. Pre-treatment spend-time and conflict nominations among the \caseMainStudents{} students of these schools define positive and negative ties. The outcome indicates whether a student wore an orange wristband during the week before the end-of-year survey, and program participants handed out these wristbands to recognize friendly or conflict-mitigating behavior. We estimate the four effects with SiDE at own invitation status zero, using known randomization probabilities. Each effect uses the students for whom every compared exposure state can occur, \caseMainNPos{} for the spend-time effect and \caseMainNInt{} for the interaction and sign-composition effects.

The estimated effect of one invited peer linked by a spend-time tie is \caseMainEstPosPP{} percentage points, about a fifth of the \vtCaseOutcomeMeanPct\% mean, with a linked-pair 95\% interval of [\caseMainLoPosPP, \caseMainHiPosPP]. After a Bonferroni adjustment across the four contrasts, this interval becomes [\vtCaseBonfLoPos, \vtCaseBonfHiPos], and we treat the analysis as exploratory. The conflict-tie estimate is $\caseMainEstNegPP$ points with interval [$\caseMainLoNegPP$, \caseMainHiNegPP], too wide to determine its direction. For two secondary outcomes, in the program and in the control schools, no estimate has an interval that excludes zero. A sign-blind comparison of untreated students with one invited peer and with none, weighted by the inverse of the known randomization probabilities, gives \vtCaseUnsignedPP{} points per invited peer. This estimate is close to the spend-time estimate, consistent with \vtCaseUnsignedSpendSharePct\% of the invited peers in this comparison being linked by spend-time ties. The signed estimates associate the positive unsigned effect with spend-time ties, which makes these peers the first to target in a confirmatory trial.

\section{Conclusion}
\label{sec:conclusion}

We defined four contrasts for peer effects in signed networks and identified each from observational data as a contrast of outcome regressions over its reachable set. Under sign-blind assignment, we showed that an unsigned increment is a weighted average of positive- and negative-tie increments, so opposing influences can cancel. We proposed SiDE, which estimates the four effects with a doubly robust score that combines sign-specific outcome models with exposure probabilities induced by individual treatment assignment, and which constructs approximate intervals that account for overlapping neighborhoods. The sign-composition effect tells a program designer whether to treat a trusted or an avoided neighbor when the number of treated neighbors is fixed. We evaluated SiDE on semi-synthetic data built on six real signed networks, and an exploratory reanalysis of a school field experiment associates a positive peer effect on wristband wearing with spend-time ties. Extending the four effects to networks that change in response to treatment is a direction for future work.

\renewcommand{\refname}{References}

\end{document}